\documentclass[twocloums]{IEEEtran}

\usepackage{cite}
\usepackage{CJKutf8}

\ifCLASSINFOpdf
 
\else
 
\fi
\usepackage{tikz}
\usepackage{pifont}
\usepackage{bbding}
\usetikzlibrary{matrix}
\usetikzlibrary{decorations.pathreplacing}
\usepackage{graphicx}
\usepackage[colorlinks,linkcolor=red]{hyperref}

\graphicspath{{images/}}
\usepackage{enumitem}
\usepackage{amsmath}
\usepackage{subfigure}
\usepackage{float}
\usepackage{cite}
\usepackage{svg}
\usepackage{amsmath}
\usepackage{graphics}
\usepackage{graphicx}
\usepackage{amssymb}
\usepackage[ruled,linesnumbered]{algorithm2e} 

\usepackage{mathrsfs}
\usepackage{siunitx}
\usepackage{booktabs}
\usepackage{bm}
\usepackage{bbm}

\usepackage{soul}
\usepackage{color, xcolor}  

\usepackage{float}
\usepackage{url}
\usepackage{amsfonts}
\usepackage[square,sort&compress,numbers]{natbib}
\usepackage{multirow}
\usepackage{textcomp}
\usepackage{supertabular}
\usepackage{array}
\usepackage{pifont}
\usepackage{hyperref}

\usepackage{soul}
\soulregister\cite7 
\soulregister\citep7 
\soulregister\citet7 
\soulregister\ref7 
\soulregister\pageref7 
\soulregister\section7
\soulregister\Function7
\soulregister\State7
\soulregister\EndFunction7
\soulregister\textbf7
\soulregister\Return7

\usepackage{algpseudocode}  
\usepackage{amsmath} 

\begin{document}

\title{Efficient Semantic Splatting for Remote Sensing Multi-view Segmentation}

\author{Zipeng Qi$^1$, Hao Chen$^2$, Haotian Zhang$^1$, Zhengxia Zou$^1$ and Zhenwei Shi$^{1,\star}$,~\IEEEmembership{Member,~IEEE} 

\vspace{6pt}
Beihang University$^1$, Shanghai Artificial Intelligence Laboratory$^2$
}


\maketitle

\begin{abstract}
Remote sensing multi-view image segmentation is a fundamental task for achieving stereoscopic perception of target scenes. This task involves processing RGB images captured from various views and producing view-consistent semantic segmentation results for each view , even under the sparse label supervision condition. Traditional training-based methods, such as CNNs and Transformers, often struggle to maintain view-consistency due to due to their limited ability to process general spatial information. Recent approaches using implicit neural networks optimize the mapping between spatial coordinates and the semantic features of sampled points, reframing the image perception task as a pixel rendering task and leveraging the spatial relationships between points to ensure view consistency and spatial continuity in the results. However, these methods typically suffer from high latency in both optimization and rendering. In this paper, we propose a novel semantic splatting approach based on Gaussian Splatting to achieve efficient and low-latency. Our method projects the RGB attributes and semantic features of point clouds onto the image plane, simultaneously rendering RGB images and semantic segmentation results. Leveraging the explicit structure of point clouds and a one-time rendering strategy, our approach significantly enhances efficiency during optimization and rendering. Additionally, we employ SAM2 to generate pseudo-labels for boundary regions, which often lack sufficient supervision, and introduce two-level aggregation losses—at the 2D feature map and 3D spatial levels—to improve the view-consistent and spatial continuity. Extensive experiments across nine datasets demonstrate the superiority of our method, achieving competitive segmentation quality with limited supervisory views. Notably, our approach reduces rendering (test) times by 90\% each while improving average mIoU by up to 3.5\%.

\end{abstract}

\begin{IEEEkeywords}
Remote sensing, Semantic segmentation, Gaussian Splatting, SAM2
\end{IEEEkeywords}

\IEEEpeerreviewmaketitle

\section{Introduction}
\label{sec:introduction}

Remote sensing image segmentation is a fundamental technology widely applied in various tasks, including change detection \cite{chen2021remote, zhang2024bifa, liu2024pixel, zhang2024cdmamba}, object extraction, e.g.,(road and building) \cite{chen2022road, zao2023topology}, and scene understanding \cite{zhu2019high, qi2020mlrsnet}. While these tasks typically produce single-view results, this paper focuses on the multi-view image segmentation task for remote sensing images, aiming to provide a more comprehensive understanding of target scenes. The inputs are RGB images captured from hundreds of views of a scene, with the outputs being segmentation results for each view. Due to the labor-intensive nature of semantic labeling, we aim to achieve multi-view segmentation using a small, easily obtainable set of annotated images — for instance, using only 3\% of labeled images from the target scene. With multi-view segmentation, we can effortlessly construct downstream outputs, such as mesh or point cloud segmentation.
\begin{figure}
    \centering
    \includegraphics[width=1.0\linewidth]{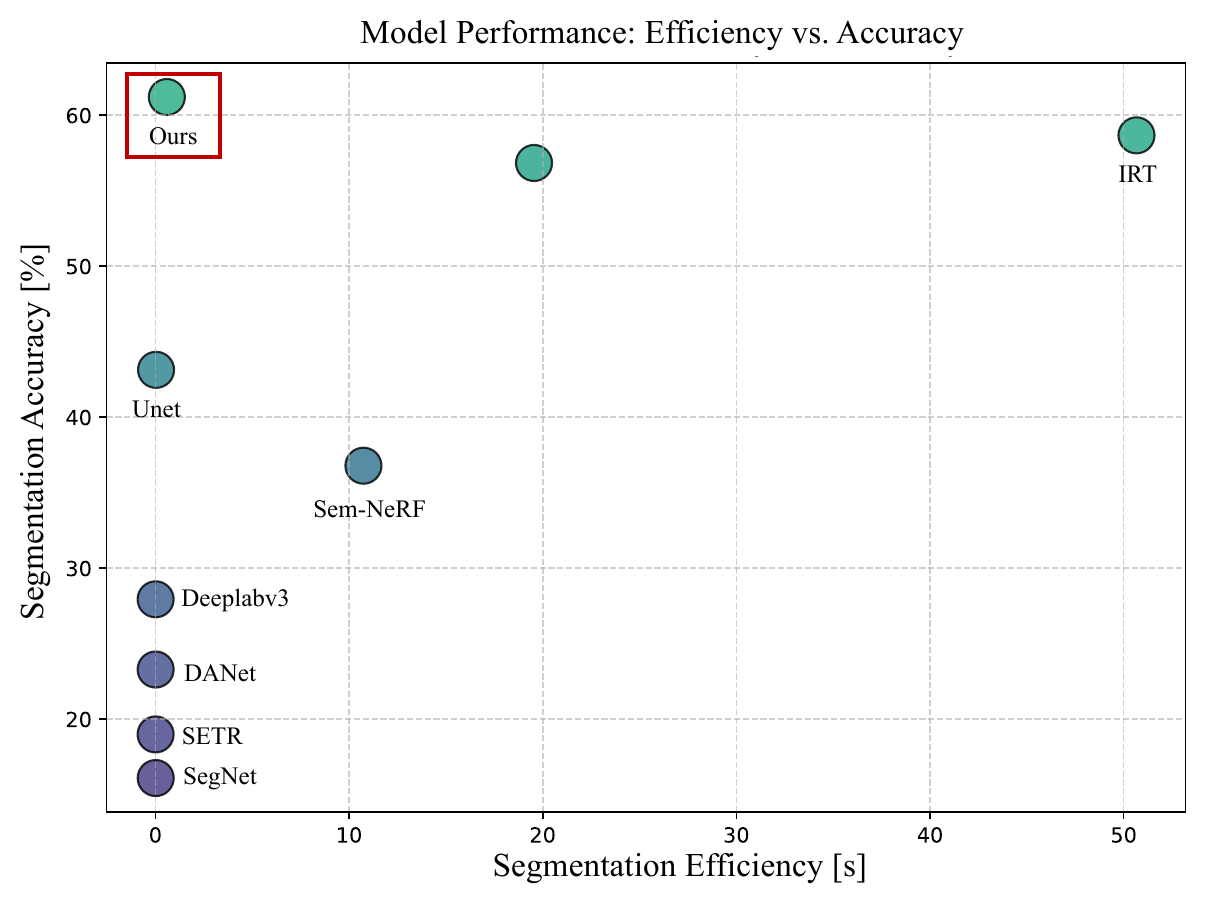}
    \caption{Our method achieves superior efficiency and accuracy for multi-view segmentation using limited labels. \textbf{Training-based} methods: SegNet, SETR, DeepLab, Uent. \textbf{Optimization-based} methods: Sem-NeRF, Color-NeRF, IRT, Ours.
    }
    \label{fig:efficencyvsaccuracy}
\end{figure}

Traditional segmentation networks, including CNN-based \cite{yuan2021multi, chen2018encoder, kayalibay2017cnn, dolz2018hyperdense} and Transformer-based \cite{zheng2021rethinking, strudel2021segmenter, huang2021missformer, li2024transformer, qi2023layered} approaches, can be directly applied to this task. However, these methods often rely on densely labeled data and extensive training resources. Moreover, they process each view independently, making it challenging to capture relationships between views or reconstruct the 3D structure of the scene, resulting in poor view consistency and low accuracy, as shown in Figure \ref{fig:efficencyvsaccuracy}. While some approaches leverage pre-trained or fine-tuned depth estimation modules to incorporate spatial information, these methods often demand even greater resources and struggle with out-of-distribution (OOD) data. Therefore, achieving multi-view segmentation with robust view consistency under sparse labeling conditions is essential to address these challenges while minimizing the reliance on exhaustive annotations and computational resources.

Recently, optimization-based segmentation methods \cite{fu2022panoptic, zhi2021place, qi2023implicit} have shown the ability to generate multi-view segmentation results using limited semantic labels. These methods divide the space covered by the views and often utilize implicit neural networks, such as NeRFs~\cite{mildenhall2021nerf, zhang2020nerf++, barron2022mip}, to map spatial positions to attributes like color and semantic information for each sampled point. Through volume rendering techniques, they can generate both color and segmentation outputs while optimizing these attributes using self-supervision. This approach implicitly constructs a continue spatial representation of the scene, thereby improving view consistency. However, the optimization process and the representation structure of implicit neural networks are often inefficient, resulting in significant optimization and rendering latencies, as illustrated in Figure \ref{fig:efficencyvsaccuracy}.

In general, training-based methods primarily focus on single-view images and require a large number of labeled data, limiting their capacity to model relationships between different views and resulting in poor performance when labels are sparse. On the other hand, optimization-based methods, while capable of achieving better view consistency, often suffer from high latency during both optimization and rendering due to their reliance on implicit representations. To overcome these challenges, we propose an efficient(typically saving 90\% testing time assumption) Gaussian Splatting-based semantic splatting method tailored for remote sensing scenes. Our approach is designed to achieve high performance(increasing average accuracy over 3.5\%) with a minimal number of labeled views (e.g., 3 labeled images per 100 views).

Unlike continuous color-related attributes that can be directly optimized, semantic classes are discrete and require a different approach. To address this, we first incorporate additional continues semantic features for each point and project these features onto the image plane to generate a semantic feature map. Subsequently, a transfer layer is introduced to derive the final segmentation results from the projected feature map. Second, the wide shooting angles in remote sensing scenes pose a challenge for accurately modeling semantic attributes, particularly in boundary regions(see Figure~\ref{fig:boundary regions}), due to the limited number of labeled views. To overcome this, we utilize foundation models such as \textit{SAM2} \cite{ravi2024sam}, which are trained on extensive datasets, to generate pseudo-labels for boundary areas. This approach improves segmentation accuracy in these challenging regions. Finally, while the basic Gaussian Splatting \cite{kerbl20233dgaussiansplattingrealtime} architecture enhances computational efficiency, it may lead to a loss of spatial continuity in attribute modeling. To address this issue, we design two aggregation losses to enforce spatial consistency at both the 2D feature map level and the 3D spatial point level, ensuring more coherent attribute representation.

We conduct extensive experiments to validate the effectiveness of the proposed method. The evaluation is performed on a multi-view segmentation dataset, which includes six sets of synthetic data generated using the well-known CARLA simulation platform ~\cite{dosovitskiy2017carla} and three sets of real-world data collected from Google Maps. Experimental results demonstrate that our method outperforms both training-based and optimization-based state-of-the-art approaches. Furthermore, visual comparisons highlight the ability of our method to deliver more accurate and view-consistent segmentation results. 

The contributions of this paper are summarized as follows:
\begin{itemize} 
    \item We propose a semantic splatting method for multi-view image segmentation under limited number of labels, firstly extending Gaussian Splatting to the field of remote sensing image semantic segmentation. Specifically, we introduce a semantic head for each sampled point and a transfer layer to generate results from the splatted feature maps. Our method achieves both optimal accuracy and efficiency;

    \item We leverage foundation models like SAM2 to generate pseudo labels for boundary regions, effectively addressing the challenge of modeling semantic attributes in these regions, which often lack supervision under limited labeled view conditions;
    
    \item We design two-level aggregation losses to enhance spatial consistency at both the 2D feature map level and 3D spatial point level, improving and spatial continue ability and segmentation performance.  
    
    \end{itemize}

The remainder of this paper is organized as follows: Section \ref{section:related work} reviews the related work. Section \ref{section:method} details the proposed method. Experimental results are presented in Section \ref{section:experiments}. Finally, conclusions are provided in Section \ref{section:conclusion}.

\section{Related Work}\label{section:related work}
To provide a clearer understanding of our method and its distinctions from existing approaches, we review the related work and preliminary of coordinate system conversion as follows:

\subsection{Training-based Segmentation}
Training-based methods have benefited significantly from the rapid advancements in convolutional neural networks (CNNs) and Transformer architectures, achieving remarkable success in various segmentation tasks. CNN-based models primarily address challenges such as preserving detailed information and extracting accurate and rich abstract features. Many of these models are built on an encoder-decoder framework, with Unet\cite{ronneberger2015unetconvolutionalnetworksbiomedical} being a prominent example. In this architecture, the encoder extracts image features through cascaded convolutional kernels while progressively downsampling the feature map resolution. The symmetrical decoder then reconstructs the feature maps and upsamples their resolution. Skip connections between the encoder and decoder are employed to retain fine-grained details during the transformation from RGB images to semantic labels. Building on this framework, dilated convolution\cite{yu2016multiscalecontextaggregationdilated} has been introduced to expand the receptive field without sacrificing resolution. This technique reduces network parameters and mitigates the risk of overfitting. Additionally, some approaches incorporate multi-scale information fusion and feature pyramid strategies\cite{ye2021enhanced, kirillov2019panopticfeaturepyramidnetworks, seferbekov2018feature, qi2024not} to capture richer abstract features and improve segmentation accuracy. Methods such as DANet\cite{fu2019dual} further enhance feature representation by leveraging dual attention mechanisms, effectively capturing both coarse object-level and fine pixel-level information to enrich feature extraction.

Recently, the Transformer architecture, initially introduced in the field of Natural Language Processing (NLP)\cite{vaswani2017attention}, has gained widespread adoption in the image processing domain for tasks such as classification, object detection, and segmentation. This architecture treats image patches as tokens and processes them through cascade layers, where the attention mechanism effectively handles long-range dependencies and explores relationships between tokens. SETR\cite{zheng2021rethinking} was the first to extend the Vision Transformer~\cite{dosovitskiy2020image} framework to image segmentation tasks. Building on this, Segformer\cite{xie2021segformer} utilizes a lightweight backbone combined with simplified MLP-like prediction heads, achieving higher accuracy while maintaining efficiency. Despite these advancements, training-based methods typically rely on large volumes of pixel-dense annotations to learn robust and generalizable features, ensuring high accuracy. However, they often struggle to model spatial relationships across multiple views, limiting their effectiveness in multi-view scenarios.

Some methods\cite{lagos2022semsegdepth, sanchez2018hybridnet} combine RGB and depth information by utilizing a dual-branch CNN backbone or incorporating self-attention mechanisms to address the limitations in modeling spatial relationships. However, these approaches rely on additional depth supervision for parameter training, which is often expensive and impractical to obtain in real-world applications. Furthermore, pre-trained depth estimators typically struggle with multi-view depth prediction, as they are usually trained on single-view datasets and fail to generalize effectively to out-of-distribution (OOD) data. In contrast, our method achieves the desired results using only RGB images and labels from a limited number of views, eliminating the need for additional depth supervision.

\subsection{Optimization-based Segmentation} 
Optimization-based methods leverage implicit neural networks, such as NeRF, which were initially designed for novel view synthesis, to process single scenes captured from multiple views. NeRF-like approaches\cite{fu2022panoptic, zhi2021place, qi2023implicit, qi2022remote, mirzaei2023spin, engelmann2024opennerf} divide 3D space by sampling spatial points along rays extending from the camera to each pixel. These methods assign color and density attributes to each sampled point using multi-layer perceptrons (MLPs). The MLPs take the position and view angle of each point as input and output the corresponding attributes, as follows:
\begin{equation}
    c_i,\sigma_i = \Phi_{d}(\Phi_{s}(x,y,z),\theta,\beta)),
\end{equation}
where $x,y,z,\theta,\beta$ are the $i_{th}$ point position and the angle of the associated ray, the $c_i$ and $\sigma_i$ are the color and density values corresponding to the $i_{th}$ point respectively, while $\Phi_{d}$ and $\Phi_{s}$ are the MLPs. Then they render final color values of each piexe with volume rendering as fellow:
\begin{equation}\label{eq:rendering}
\mathrm{Color}= \sum \limits ^{N}_{i=1} \exp{\Big(-\sum \limits ^{i-1}_{j=1}\alpha_j\sigma_j\Big)}\Big(1-\exp{(-\alpha_i\sigma_i})\Big)c_i,
\end{equation}
where $\alpha_i$ is the interval distance between the $i$ point and the $i+1$ point along the corresponding ray. They compare the rendered results and captured image of each view within a self-supervision to optimize the parameters of MLPs. The optimized MLPs establish a query relationship between positions and attributes. Accurate query mapping enables effective rendering of images from novel views. 

Some methods extend the rendering process from RGB images to semantic labels. For example, Sem-NeRF\cite{zhi2021place} introduces an additional branch to generate semantic features $s_i$ for each sampled point. It simultaneously renders semantic labels and RGB images using a volume rendering approach. However, Sem-NeRF determines $s_i$ solely based on positional information. Building on this, Color-NeRF\cite{qi2022remote} incorporates color information into the rendering process for semantic features, enhancing the segmentation of small objects. In remote sensing scenes, the large pitch angles often result in a significant number of sampling points in the air, which can affect rendering accuracy. To address this challenge, IRT\cite{qi2023implicit} introduces a selection layer to adjust the sampling and rendering processes. Furthermore, it enriches the integration of color information into semantic feature rendering through a lightweight texture extraction network. It is also noteworthy that NeRF-like methods naturally support the extraction of depth maps or mesh representations using density attributes. Compared to traditional approaches such as TSDF\cite{qi20223d} or point clouds \cite{guo2020deep}, implicit spatial representations offer distinct advantages, including a smaller memory footprint and spatial resolution independence.

With the development of Gaussian Splatting, several methods \cite{shen2025flashsplat, ye2025gaussian, choi2025click} have extended its application from RGB rendering to segmentation tasks. However, these approaches predominantly focus on indoor scenes and rely on all-view supervision to achieve instance identification. In contrast, our work targets remote sensing scenes and emphasizes semantic segmentation, particularly under sparse-view supervision. The basic preliminary of Gaussian Splatting will be introduced in Section \ref{sec:CR}.

\subsection{Segment Anything}
Segment Anything (SAM) \cite{kirillov2023segment} is a foundational vision model designed to generate accuracy image segmentation based on input visual prompts, such as points or bounding boxes. The model consists of three key modules: an image encoder, a prompt encoder, and a mask decoder. The masks produced by SAM represent regions that may contain the target objects. Depending on the number and positioning of the prompts, SAM generates potential segmentation results at multiple levels. Trained on a large volume of data, SAM demonstrates robust performance and excels across a wide range of segmentation tasks\cite{ke2024segment, mazurowski2023segment, cen2023segment, qi2024multi}. SAM2 \cite{ravi2024sam} builds upon SAM, extending its capabilities from the image domain to the video domain. SAM2 can accept various types of prompts, including clicks (positive or negative), bounding boxes, or masks, to define the extent of an object within a specific video frame. The lightweight mask decoder then outputs a segmentation mask for the current frame based on the image embedding and encoded prompts. In the video setting, SAM2 propagates the mask prediction across all frames. Additional prompts can be iteratively provided in subsequent frames to further refine the segmentation. To ensure accurate mask predictions across all frames, SAM2 incorporates a memory mechanism, consisting of a memory encoder, memory bank, and memory attention module.

However, both SAM and SAM2 are limited to generating non-semantic masks, which are not directly applicable to remote sensing images. Fortunately, these non-semantic masks are adequate for our approach. We use SAM2 to generate pseudo-labels for boundary regions to improve the segmentation results.

\subsection{Coordinate System Conversion}
\label{sec:CSC}
Our method involves the conversion of four coordinate systems including the pixel coordinate system, image coordinate system, camera coordinate system, and world coordinate system. Therefore, in this subsection, we introduce transformations between different coordinate systems.

The conversion relationship between the pixel coordinate system and the image coordinate system is defined as follows: 
\begin{equation}
    \begin{bmatrix}
       u \\ v \\ 1 
   \end{bmatrix} = 
  \begin{bmatrix}
  \frac{1}{d_x}&0&u_0\\
  0&\frac{1}{d_y}&v_0\\
  0&0&1
  \end{bmatrix}
   \begin{bmatrix}
       x \\ y \\ 1 
   \end{bmatrix},
\end{equation}
where $x,y$ and $u,v$ represent the pixel and image coordinates respectively, $d_x$ and $d_y$ are the width and height of a pixel corresponding to the photosensitive point. The camera coordinate system to the image coordinate system follows a perspective transformation relationship, which can be represented by using similar triangles:
\begin{equation}
    z_c    
    \begin{bmatrix}
       x \\ y \\ 1 
   \end{bmatrix} =
   \begin{bmatrix}
  f&0&0&0\\
  0&f&0&0\\
  0&0&f&0
  \end{bmatrix}
   \begin{bmatrix}
       x_c \\ y_c \\ z_c\\1 
   \end{bmatrix} =
  \begin{bmatrix}
    \mathbf{K}|\mathbf{0}  
  \end{bmatrix}
   \begin{bmatrix}
       x_c \\ y_c \\ z_c\\1 
   \end{bmatrix},
\end{equation}
where $x_c,y_c,z_c$ are the camera coordinate, $f$ is the focal length and $K$ is the intrinsic parameter matrix of the camera. The world coordinate system can be obtained from the camera coordinate system through rotation and translation:
\begin{equation}
    \begin{bmatrix}
       x_c \\ y_c \\ z_c\\1 
   \end{bmatrix} =
       \begin{bmatrix}
       \mathbf{R} & \mathbf{t} \\ 
       \mathbf{0}_{1\times 3}&1 
   \end{bmatrix}
   \begin{bmatrix}
       x_w \\ y_w \\ z_w\\1 
   \end{bmatrix}
\end{equation}
where $R$ represents the rotation matrix and $t$ the translation matrix, with $[R|t]$ being the camera's extrinsic parameters. The coordinates $x_w$, $y_w$, and $z_w$ refer to the absolute positions of objects in the world coordinate system. In our work, we employ COLMAP~\cite{schonberger2016structure} to estimate the intrinsic and extrinsic parameters for each viewpoint.

\begin{figure*}[t]
\centering
\includegraphics[width=\linewidth]{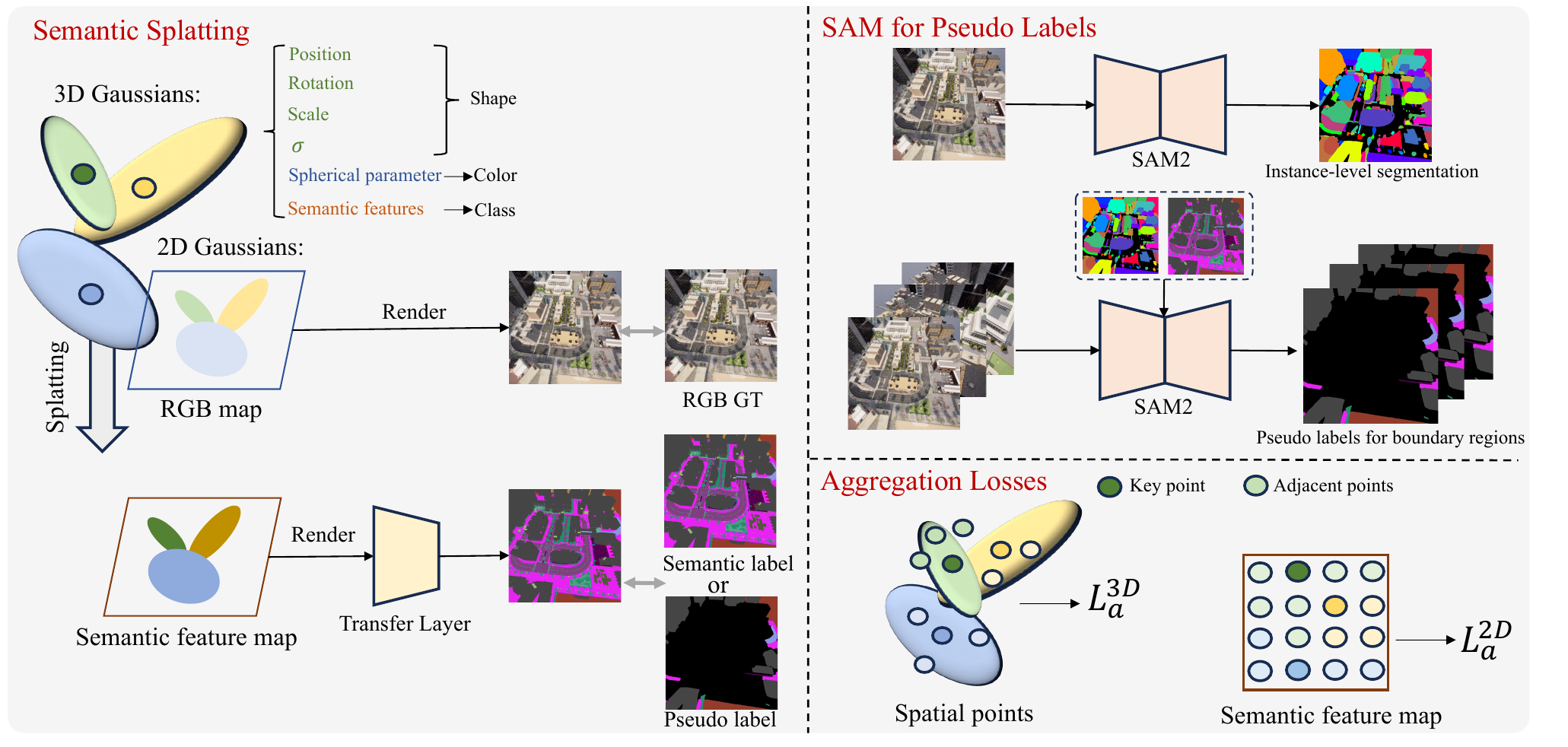}
\caption{The overview of the proposed model. Our method is an optimization-based semantic splatting approach for multi-view segmentation in remote sensing. Leveraging an explicit point cloud structure and volume rendering, it achieves high accuracy and low latency in generating RGB and semantic segmentation results(see Section \ref{sec:CR} and \ref{sec:SR}). Aggregation losses improve spatial generalization(see Section \ref{sec:AL}), while SAM2 generates pseudo-labels for boundary regions, further enhancing boundary segmentation quality (see Section \ref{sec:SPL}).}
\label{method}
\end{figure*}

\section{Methodology}\label{section:method}
An overview of our method is illustrated in Figure \ref{method}. We propose an optimization-based, efficient semantic splatting approach for multi-view segmentation in remote sensing. The input to our method consists of a set of RGB images captured from various views, along with a limited number of labeled images. The output is the semantic segmentation results for the remaining views or any novel views. Unlike the implicit neural networks, our method employs an explicit point cloud structure. The color-related attributes and semantic feature attributes of each point are splatted onto the image plane. In the splatting process, we then apply volume rendering to generate both RGB and semantic segmentation results simultaneously(see Section \ref{sec:CR} and \ref{sec:SR}). Similar to other optimization-based segmentation methods, the attributes are optimized in a self-supervised manner. The explicit structure enhances the efficiency of optimization and rendering. However, it has weaker spatial generalization capabilities compared to implicit neural networks, particularly in boundary regions. To overcome this limitation, we introduce aggregation losses at both the 2D feature map and 3D spatial point levels(see Section \ref{sec:AL}). Additionally, we leverage SAM2 to generate pseudo-labels for the boundary areas of each view(see Section \ref{sec:SPL}). The detailed inference process of our method is shown in Algorithm \ref{alg:algorithm1}.


\begin{algorithm}[t] 
\label{alg:algorithm1}
\caption{Semantic Splatting}
\KwIn{$\{(\mathbf{I}_n, \mathbf{C}_n, \mathbf{L}_m)|n = 1 : N, m = 1:M\}$ ($N$ images($I$) with camera parameters($C$) and $M$ labels ($L$), $M \ll N$), and $k$ initialized 3D Gaussians with color and semantic attributes.}
\KwIn{$\mathbf{T}$ (iterate number) and $\{\mathbf{P_j}|j = 1:N-M\}$ (pseudo labels for other views generated by SAM2)}
\KwOut{Optimized attributes of 3D Gaussians and $\mathbf{S}^{tgt}$ (multi-view segmentation)} 
\BlankLine
//optimization\\
\For{$i$ in $1:\mathbf{T}$}
{
// select a view image from $I_n$\\
$\mathrm{I_i},\mathrm{C_i} = \mathrm{Sample}(\mathrm{I_n}, \mathrm{C_n})$\\
// Splatting 3D Gaussians into the $i_{th}$ image plane to get 2D Gaussians with color and semantic attributes\\
C, S = Splatting($G_1,G2,\dots G_k | C_i$) \\
//render color and semantic results\\
Color, Label = R(C, S) //Eq. \ref{eq:color-render} and Eq.\ref{eq:label-render}\\
// Calculate Loss\\
$\mathbf{L} = Loss(\{Color, I_i\}, \{Label, L_i/P_i\})$ // Eq. \ref{eq:loss} \\
//gradient descent to optimize parameters of the 3D Gaussians
}
\BlankLine
//inference\\
\quad //render segmentation result from novel view\\
$\mathbf{S}^{tgt}$ = Splatting$(G_1,G2,\dots G_k | C)$
\end{algorithm}

\subsection{Color Rendering}
\label{sec:CR}
In this subsection, we first introduce the splatting technique for rendering, which forms the foundation of our method for color and semantic rendering. Splatting, also known as the footprint method, differs from ray casting by repeatedly computing the projection and accumulation effects of spatial positions, rather than calculating direct light paths. It uses a footprint function to determine the influence range of each spatial point. By modeling the intensity distribution of points or small pixel areas with a Gaussian distribution, splatting calculates their overall contribution to the image. These contributions are then synthesized to produce the final rendered result. Due to its low latency in the rendering process, splatting is widely used in graphics applications \cite{niedermayr2024application, palandra2024gsedit, chen2024text}.

Gaussian Splatting\cite{kerbl20233dgaussiansplattingrealtime} extends splatting technology to the field of novel view synthesis. It typically uses COLMAP to initialize a point cloud that represents the spatial information. Each point in the point cloud is assigned a 3D Gaussian function. The shape of each 3D Gaussian function is defined by its mean $\mu^{3D}$, representing the center position $(x,y,z)$, and its covariance $\Sigma^{3D} = RSS^TR^T$, which defines the radiance field. Here, $S$ and $R$ denote the scaling and rotation vectors of the Gaussian ellipsoid, respectively. For the color attribute, spherical functions, are employed to output the color vector $c$ based on the input rendering angle, capturing the view-dependent appearance of the scene. The density attribute $\sigma$is used to represent opacity. During the splatting process, the 3D Gaussians are converted into 2D Gaussians via coordinate system transformation (see Section \ref{sec:CSC}).
For a given pixel position, the view transformation computes the distances to all overlapping point scopes, representing their respective depths. This results in a sorted list $N$ of valid points. Subsequently, alpha blending—a technique for mixed alpha synthesis—is applied to calculate the final color of the overall image. One pixel color alpha blending process is shown as follows:
\begin{equation}
    C^{one} = \sum_{i \in N} c_i \alpha'_i \prod_{j=1}^{i-1} (1 - \alpha'_j),
    \label{eq:color-render}
\end{equation}
where $C$ is the render color vector, $c_i$ is the color attribute vector, and the final opacity $\alpha_i$ is the optimized opacity related to the opacity $\alpha_i$
\begin{equation}
    \alpha'_i = \sigma_i \exp\left(-\frac{1}{2}(p_i - \mu_i^{2D})^T (\Sigma_i^{2D})^{-1}(p_i - \mu_i^{2D})\right),
\end{equation}
where $p_i$ is the pixel position, $\mu_i^{2D}$ and $\Sigma_i^{2D}$ are the mean vector and covariance matrix of the $i_{th}$ 2D Gaussian. Typically, the $\mathcal{L}_{1}$ and $\mathcal{L}_{D-SSIM}$ are used to calculate the loss and then optimize the parameters in a self-supervision mechanism. The accuracy of color rendering results plays an important role in scene spital structure estimation, which is also an basic information for smantic splatting.

\subsection{Semantic Rendering}
\label{sec:SR}
In this subsection, we introduce semantic rendering. Unlike color attributes, which exist in a continuous domain, semantic classes are discrete values, making the direct optimization of semantic classes for each point a non-differentiable problem. To overcome this, we transform the non-differentiable optimization into a differentiable process. Specifically, we elevate the one-dimensional discrete class values into high-dimensional continuous semantic features $s$ and assign these features to each point, as illustrated in Figure \ref{method}. Subsequently, we render one pixel semantic feature maps using alpha blending:
\begin{equation}
    S^{one}_f = \sum_{i \in N} s_i \alpha'_i \prod_{j=1}^{i-1} (1 - \alpha'_j).
    \label{eq:label-render}
\end{equation}
The point shape information including $\mu^{3D}$, $\Sigma^{3D}$ and $\alpha$ as well as the splatting process, is shared between color and semantic rendering. Here, $s_i$ represents the semantic attribute of the $i_{th}$ point and $S_f$ denotes the rendered semantic features corresponding to one pixel. Finally, a decoder composed of simple MLPs is applied to generate semantic segmentation from the whole semantic feature maps(composed from all pixel semantic features):
\begin{equation}
    S^{all} = \mathrm{Decoder}(S^{all}_f).
\end{equation}
Replacing discrete class values with continuous semantic features offers two key advantages:
(1) \textit{Differentiability}: This transformation converts non-differentiable problems into differentiable ones, enabling smoother optimization, albeit with a slight reduction in efficiency.
(2) \textit{Enhanced Expressiveness}: This approach increases the model’s ability to represent semantic attributes in complex scenes, making it more effective at handling intricate segmentation tasks.
We set the cross-entropy loss to calculate semantic loss $\mathcal{L}^g_s$ for rendered semantic results $S^{all}$ ground truth $S^{GT}$.

\subsection{SAM for Pseudo Labels} 
\label{sec:SPL}
\begin{figure}
    \centering
    \includegraphics[width=0.9\linewidth]{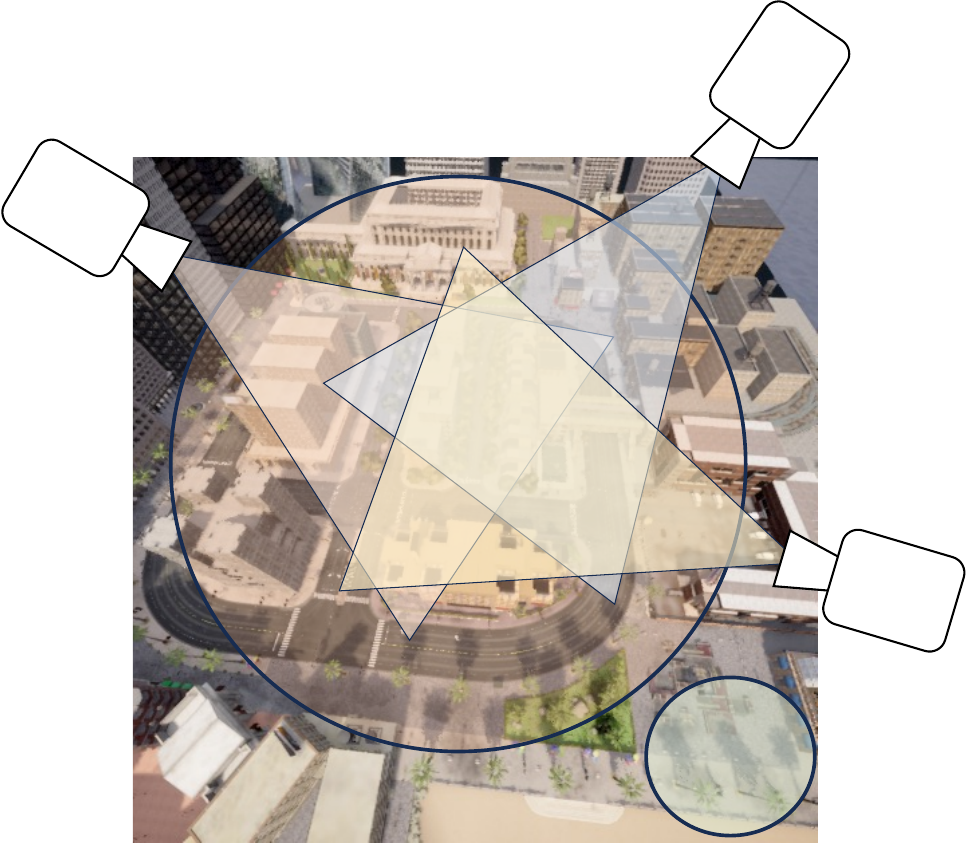}
    \caption{For a target scene with only a few views, e.g., 3 views with supervision labels, some regions will lack supervision. \textbf{Large circle}: The common region covered by supervised views. \textbf{Small circle}: The boundary regions lacking supervision.}
    \label{fig:boundary regions}
\end{figure}

Challenges in scene segmentation often emerge in boundary regions, as these areas may extend beyond the common field of view of the supervisory views, leading to insufficient monitoring—particularly when only a limited number of supervisory views are available. This is clearly illustrated in Figure \ref{fig:boundary regions} . While the Gaussian-like splatting architecture demonstrates notable rendering efficiency, its limited spatial continuity exacerbates the segmentation difficulty under our sparse label setting. To mitigate this challenge, we employ SAM2, a widely used zero-shot image/video segmentation method, to generate pseudo labels for boundary regions.
However, the results from SAM2 lack rich semantic information, making them unsuitable for direct use in semantic segmentation tasks. 

To address this, we first randomly select a supervisory view and input the corresponding RGB image into SAM2 to generate a single-view full-segmentation result, treating regions that may correspond to objects as individual instances. From this result, we generate a boundary mask $B$, where $0$ represents non-boundary areas and $1$ denotes boundary-intersecting areas. Next, we assign each instance a pseudo-class based on the label from the selected view, determining the class with the maximum number of pixels within the corresponding area. Finally, we use each instance mask as a visual prompt and re-input it into SAM2 to generate multi-view segmentation results for each instance, assigning the same pseudo-class label to all corresponding segmentation results, noted as $S^p$. 
We set the cross-entropy loss $\mathcal{L}^p_s$ for views with pseudo labels as fellowing:
\begin{equation}
\mathcal{L}_{s}^p = CE(S^{all} * B, S^{p} * B)
\end{equation}
Here, $B$ is the boudary region mask.

\subsection{Aggregation Loss}
\label{sec:AL}
The limited number of labeled views poses a significant challenge, as it hinders the optimization of semantic attributes for certain points in specific regions. Although the explicit representation used in our method provides notable advantages in optimization and rendering efficiency, its ability to continuously represent spatial information is weaker compared to implicit neural networks. This limitation affects the model's capacity to generalize effectively in unobserved areas. The primary cause of this limitation lies in the inherent differences in data structures. Implicit neural networks rely on MLPs to model the relationships between spatial locations and attributes, allowing them to be more robust to nearby spatial variations and enabling smooth transitions across unobserved regions. In contrast, our method utilizes an explicit point cloud structure, where spatial generalization is inherently constrained.

To address this challenge, we propose two levels of aggregation losses: one for the 2D rendered feature map and the other for the 3D spatial semantic features. The 2D aggregation loss is designed to enhance the similarity of features from adjacent points in the rendered 2D feature map, while the 3D aggregation loss focuses on increasing the similarity of semantic features from neighboring points in 3D space. Specifically, for the 2D aggregation loss, we randomly select $m$ feature points from the rendered feature map and calculate the mean similarity between each selected point and its $k$, e.g., $k=5$ nearest neighbors, based on their 2D pixel coordinates. 
\begin{equation}
    \mathcal{L}_a^{2D} = \frac{1}{mk}\sum^m_{i=1}\sum^k_{j=1}s_i\log(\frac{s_i}{s_j}).
\end{equation}
For the 3D aggregation loss $\mathcal{L}_a^{3D}$, we adopt a similar approach. The key distinction lies in the selection process: $m$ points are randomly chosen from the point cloud, and their 3D coordinates are used to identify the $k$ nearest neighboring points.

\subsection{Optimization Process, Details and Total Loss}
\textbf{Optimization process.} For each optimization step, we select a view and obtain the rendered RGB and segmentation results. The captured images, along with their corresponding semantic labels or pseudo-labels, are used to optimize the parameters of the corresponding points in the point cloud. Given the inherent inaccuracies in pseudo-labels and their overwhelming proportion compared to ground-truth labels (e.g., $3\%:97\%$), we manually set the ratio of views with ground-truth labels to pseudo-labeled views as $1:8$. The ratio is important for the performance, the parameter experiments show the related results.
We use COLMAP to extract the structure-from-motion (SfM) of the scene as the initial point cloud. The spatial structure of the point cloud is crucial for ensuring the accuracy of the rendered results, including both RGB images and segmentation outputs. Therefore, we employ split and merge strategies every 2000 steps for both the RGB attributes and semantic features of the points, adjusting them according to the gradient. This process is similar to Villain Gaussian Splatting.

\textbf{Loss function.} We use $\mathcal{L}_1$ and $\mathcal{L}_{D-SSIM}$ for optimizing color-related parameters. We define $\mathcal{L}_s^g$  as $\mathcal{L}_s$  when selected view contains ground truth label, and  $\mathcal{L}_s^p$ as $\mathcal{L}_s$ when selected view contains pseudo label. The $\mathcal{L}_s$ is used for optimizing semantic-related parameters. Additionally we employ $\mathcal{L}_a^{2D}$ and $\mathcal{L}_a^{3D}$ to increase the similarity between 2D feature points and 3D spatial points. The total loss is the weighted above losses:
\begin{equation}
    \mathcal{L} = \mathcal{L}_1 + \mathcal{L}_{D-SSIM} + \mathcal{L}_s + a*\mathcal{L}_a^{2D} + b * \mathcal{L}_a^{3D}.
    \label{eq:loss}
\end{equation}
Here, $a$ and $b$ is the hyper-coefficient for aggregation losses.

\textbf{Details.} We provide the parameter details of our method. The optimization process consists of 30,000 steps, with a maximum of 300,000 points in the point cloud. The number of semantic feature channels is set to 16, and the transfer layer maps these 16 channels to the number of semantic classes. For both $\mathcal{L}_a^{2D}$ and $\mathcal{L}_a^{3D}$, the number of neighboring points is set to $k=5$ . The candidate coefficient for $\mathcal{L}_a^{2D}$ and $\mathcal{L}_a^{3D}$ are 1.0,0.5 and 0.1 respectively. The choice of coefficient for each sub-loss is crucial to performance, as demonstrated by our parameter experiments.

\section{Experimental Results and Analysis}\label{section:experiments}

\subsection{Dataset and Metrics}
\textbf{Dataset.} In this paper, we evaluate our method using the challenging dataset constructed in IRT \cite{qi2023implicit}. This dataset comprises six synthetic sub-datasets generated using the CARLA platform and three real sub-datasets sourced from Google Maps. Each scene in the dataset is captured by extracting key frames from videos at equal intervals, resulting in approximately 100 images per scene taken from various shooting angles. The dataset includes a total of 865 images, each with a resolution of $512 \times 512$ pixels. As shown in Table~\ref{datasetdescription}, only 2\% to 6\% of the images in the training set are annotated with corresponding labels. The dataset covers scenes of varying scales, ranging from individual buildings to entire towns. Annotations for the synthetic data are generated by CARLA and include 20 types of ground objects such as buildings, roads, zebra crossings, and vegetation. For the real datasets, more than 7 common ground object types are labeled, including buildings, roads, and vehicles.

\begin{table}[!htb] \small
\centering
\caption{Details of the synthetic and real dataset.}
\label{datasetdescription}.
\begin{tabular}{c|cccc}
\toprule
         & \#views & labeling ratio & \#classes \\ 
\midrule
sys \#1 & 100   & 3\%  & 20     \\ 
sys \#2 & 100   & 4\%  & 18    \\ 
sys \#3 & 100   & 5\%  & 20    \\ 
sys \#4 & 80    & 6\%   & 19  \\  
sys \#5 & 85    & 6\%   & 19  \\  
sys \#6 & 100   & 5\%   & 18  \\  
real \#1 & 100  & 2\%   & 5 &  \\  
real \#2 & 100  & 2\%   & 3 &  \\  
real \#3 & 100  & 2\%   & 3 & \\  
\bottomrule
\end{tabular}
\end{table}

\begin{figure*}[!htb]
\centering
\includegraphics[width=\linewidth]{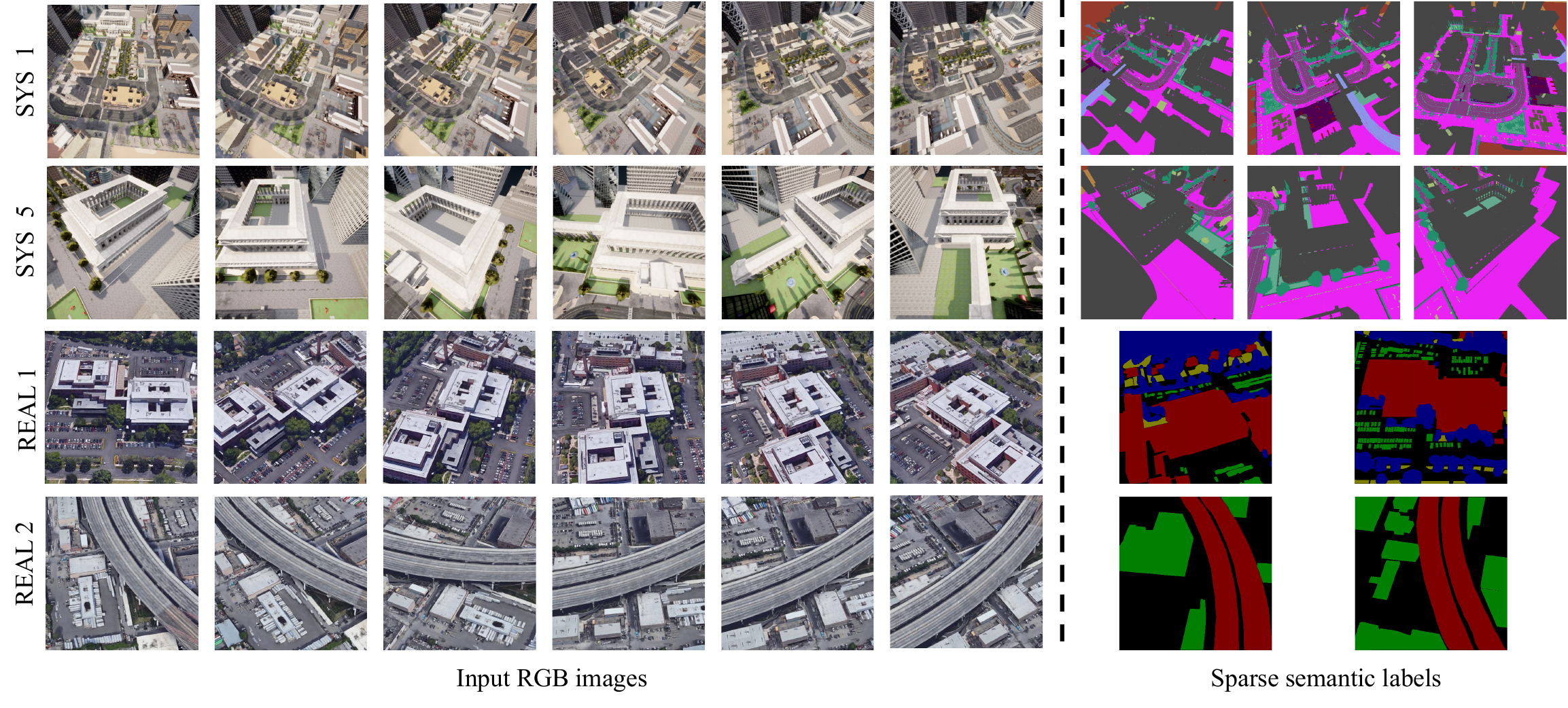}
\caption{ We present input samples of four sub-datasets. \textbf{Left}: Six sampled input RGB images. \textbf{Right}: Sparse semantic labels, with real scenes having only two labeled views. The type includes from area level (SYS 1 and REAL 1) to local building level(SYS5 and REAL 2).}
\label{datasample}
\end{figure*}

\textbf{Metric.} In our experiment, we employ the mean Intersection over Union (mIoU), a widely used metric for segmentation tasks, to evaluate and compare the performance of different methods.
\begin{equation}
    MIoU = \frac{TP}{FP+FN+TP}
\end{equation}
where the $TP$ is the true positive, $FP$ is the false positive, $FN$ is the false negative.

\subsection{Comparison Methods}

To validate the effectiveness of our method, we selected several methods for comparative experiments. We compared our approach with training-based segmentation methods, including CNN-based techniques such as SegNet \cite{badrinarayanan2017segnet}, Unet \cite{ronneberger2015u}, DANet \cite{fu2019dual}, and DeepLabv3 \cite{chen2017rethinking}, as well as the Transformer-based SETR \cite{zheng2021rethinking}. Additionally, we primarily focused on comparisons with optimization-based methods. Sem-NeRF \cite{zhi2021place} performs joint color reconstruction and semantic segmentation by utilizing two MLP heads to process spatial features, predicting RGB values and semantic labels simultaneously. The baseline strategy employs a uniform renderer function for rendering both color results and semantic segmentation. Another two-stage strategy involves first reconstructing color information and then converting it into segmentation results \cite{qi2022remote}. IRT \cite{qi2023implicit} enhances NeRF-based methods by incorporating CNN features as additional tokens to improve the processing of complex textures. The comparison methods are summarized as follows:

\begin{enumerate} 
\item SegNet \cite{badrinarayanan2017segnet}: A CNN-based semantic segmentation method utilizing an encoder-decoder architecture.
\item Unet \cite{ronneberger2015u}: A CNN-based semantic segmentation method employing skip connections to preserve detailed information from encoder layers.
\item DANet \cite{fu2019dual}: A CNN-based semantic segmentation method introducing a dual attention mechanism to adaptively integrate local features and global dependencies.
\item DeepLabv3 \cite{chen2017rethinking}: A CNN-based semantic segmentation method leveraging dilated convolutions to effectively enlarge the receptive field.
\item SETR \cite{zheng2021rethinking}: A Transformer-based semantic segmentation method.
\item Sem-NeRF \cite{zhi2021place}: The first NeRF-based model designed for indoor image semantic segmentation.
\item Color-NeRF~\cite{qi2022remote}: A NeRF-based method for semantic segmentation that introduces an additional color-radiance network to fuse pixel-level color information, improving segmentation performance.
\item IRT \cite{qi2023implicit}: A NeRF-based method combining ray features with CNN features to enhance segmentation accuracy.
\end{enumerate}

\subsection{Overall Comparison Results}

\begin{figure*}[!htb]
\centering
\includegraphics[width=\linewidth]{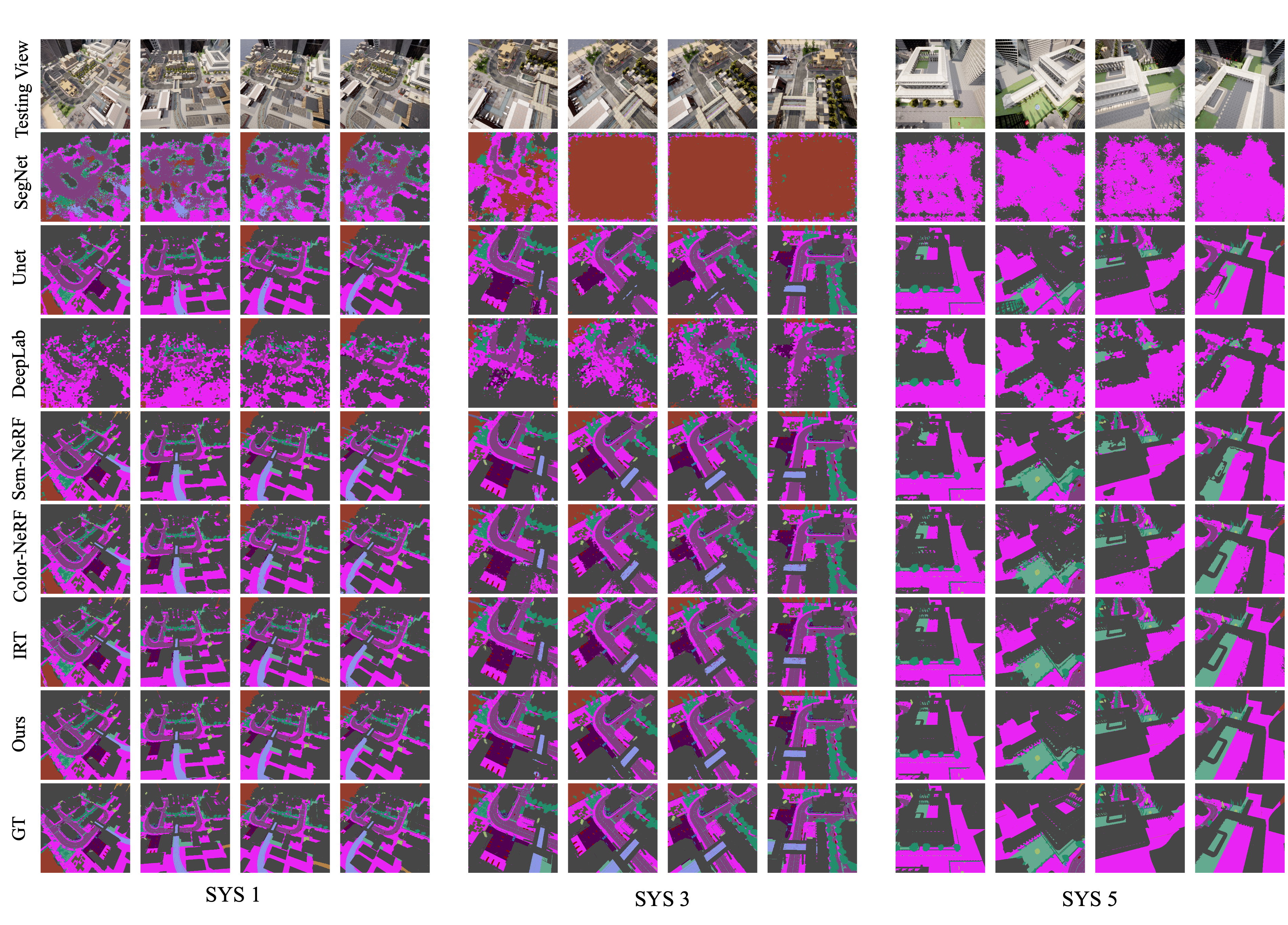}
\caption{The visual results from sys \#1, sys \#3 and sys \#5. The results show that our method achieves more accuracy and outperform other training-based and optimization-based methods in multi-view segmentation taks for both complex synthesis and real scenes.}
\label{fig:visual results1}
\end{figure*}

\begin{figure*}[!htb]
\centering
\includegraphics[width=\linewidth]{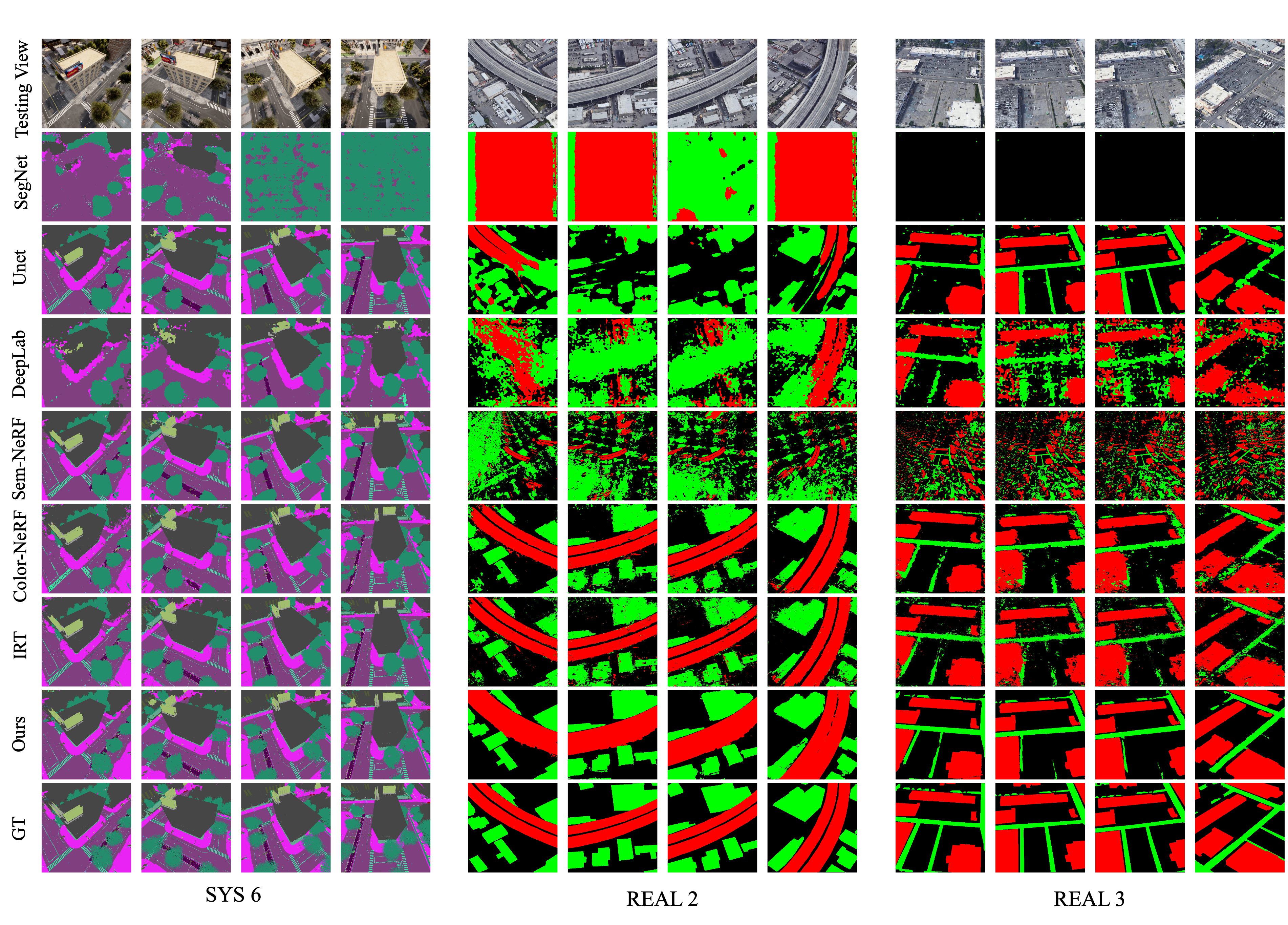}
\caption{The visual results from sys \#6, real \#2 and real \#3. The results show that our method achieves more accuracy and outperform other training-based and optimization-based methods in multi-view segmentation taks for both complex synthesis and real scenes.}

\label{fig:visual results2}
\end{figure*}

\textbf{Quantitative Analysis}
\begin{table*}[!htb] 
\centering
\caption{mIoU metric and testing time of different methods on each sub-dataset.}
\label{metric-results}
\begin{tabular}{cc|ccccccccccc}
\toprule
\multicolumn{2}{c|}{Method}& sys \#1 & sys \#2 & sys \#3 & sys \#4 & sys \#5 & sys \#6 & real \#1 & real \#2 & real \#3 & AVG & Time \\\midrule
\multirow{5}{*}{Training-based}
&SegNet &11.79& 13.21 & 10.81 & 26.71 & 8.21 & 18.77 & 8.67 & 27.88 & 18.84 & 16.10 & 0.016s\\ 
 &Unet &23.92  & 31.73 &42.26 & 41.79 & 26.63 & 38.72& 64.94 & 49.95& 68.33& 43.14 & 0.034s  \\ 
&DANet  & 9.73 & 13.91 & 16.62 &  26.78 & 8.53 & 15.79 & 35.71& 49.13& 34.30&  23.39  & 0.013s \\ 
 &Deeplab  & 18.89 & 16.64 & 19.96 & 30.42 & 11.90& 20.52 & 39.35 & 41.37& 52.37 & 27.94 & 0.012s   \\ 
 &SETR & 10.34&10.71 &12.45 &21.13& 8.90 & 13.97& 36.26 & 31.04 & 26.07 & 18.99 & 0.011s \\ \midrule
 \multirow{3}{*}{Optimization-based}
 &Sem-NeRF  & 55.73 & 34.81& 49.82 &57.24& 41.44 & 41.85 & 11.64 & 18.78& 19.76 &36.79& 10.743s \\ 
 &Color-NeRF & 58.03 &38.46 & 50.86 & 59.14 & 43.77 & 43.53 & 62.04 & 85.85 & 69.92 & 56.84&19.552s \\ 
&IRT & 57.86 & \textbf{43.23} & 53.47 & 59.98 & 43.73 & 48.59 & \textbf{65.84}  & 84.31 & 71.02 & 58.67 & 50.662s\\  \midrule
Optimization-based &Ours &\textbf{61.45} &38.98 &\textbf{58.51} & \textbf{62.99} & \textbf{46.40} & \textbf{52.14} & 63.31 & \textbf{86.87} & \textbf{80.21} & \textbf{61.21} & \textbf{0.591s} \\
\bottomrule

\end{tabular}
\end{table*}
From Table~\ref{metric-results}, it is evident that all optimization-based methods outperform training-based methods. This is primarily because training-based methods rely on learning the data distribution from large datasets to construct stable and generalized models. As a result, these methods struggle in tasks with limited data, particularly when only 3\% of the input images are labeled. In contrast, optimization-based methods typically optimize a mapping from spatial point positions and associated attributes, including both color and semantic features, for each target scene. During the optimization process, the network can learn the 3D information of the scene, ensuring view consistency in the results. When comparing Sem-NeRF, Color-NeRF, and IRT, the continuous integration of additional RGB information during semantic rendering improves segmentation accuracy. However, these methods face challenges related to high latency in both optimization and rendering, which limits their practical applicability.

Our method achieves higher mIoU scores than all the compared methods across all sub-datasets, including both synthetic and real data. Specifically, compared to Sem-NeRF, our method outperforms it by an average of 24.42\% in mIoU. We also surpass Color-NeRF by 3.54\% on average mIoU. Notably, our method outperforms IRT, the current state-of-the-art method, by over 15.5\% in average mIoU. Furthermore, our method exhibits significantly lower latency in both optimization and rendering compared to other optimization-based methods. For instance, our method saves 90\% testing time compared to IRT. This combination of high efficiency and accuracy makes our method more viable for practical applications.

We assessed the testing time of various methods. While training-based methods exhibit remarkable efficiency, their accuracy is insufficient for real-world applications. Conversely, optimization-based methods like Sem-NeRF, Color-NeRF, and IRT deliver higher accuracy but at the cost of extended computation times, requiring hours for optimization and several seconds to render results for each view. In contrast, our method achieves an excellent balance between accuracy and efficiency. It combines high accuracy with drastically reduced latency, completing rendering in less one second and requiring only about 10 minutes for optimization. This remarkable combination makes our approach significantly more practical and suitable for real-world applications.

\subsection{Qualitative Analysis}
\textbf{Visually Analysis} We present results from six testing views across four selected synthetic sub-datasets and two real sub-datasets in Figure~\ref{fig:visual results1} and~\ref{fig:visual results2}. For each sub-dataset, we display the input RGB images alongside their corresponding semantic segmentation results. The comparison includes three training-based methods—SegNet, Unet, and DeepLab—and three optimization-based methods—Sem-NeRF, Color-NeRF, and IRT. 

From the results, it is evident that optimization-based methods consistently outperform training-based methods, as discussed in the quantitative analysis section. Under sparse label conditions, training-based methods struggle to develop robust models capable of distinguishing complex background textures from object textures, often resulting in misclassification within background regions. For example, SegNet and DeepLab exhibit poor performance in background modeling for sys \#1 and sys \#6. Additionally, training-based methods demonstrate significant limitations in segmenting small objects, as seen in the segmentation of small trees and zebra crossings in sys \#1, as well as small roads in real \#3, further highlighting their weaknesses in such scenarios. Furthermore, the visual results reveal poor view-consistency in training-based methods. This is primarily due to their inability to model spatial relationships effectively across different views, leading to inconsistent predictions and degraded performance in multi-view segmentation tasks.

Optimization-based methods transform the segmentation task into a rendering process, producing class predictions for each pixel. This rendering approach offers distinct advantages: it achieves higher segmentation accuracy, even with sparse labels, and performs well in segmenting small objects. The visual results clearly demonstrate these benefits. Notably, our method delivers highly detailed and accurate segmentation results, as shown in roads in real\#3 and building details in sys \#5. 
However, other optimization-based methods, such as Sem-NeRF, Color-NeRF, and IRT, often incur significant latency during processing. In contrast, our approach leverages a splatting strategy to enhance efficiency. This remarkable combination of accuracy and efficiency positions our method as a promising solution for practical applications.

\begin{figure*}[!htb]
\centering
\includegraphics[width=0.8\linewidth]{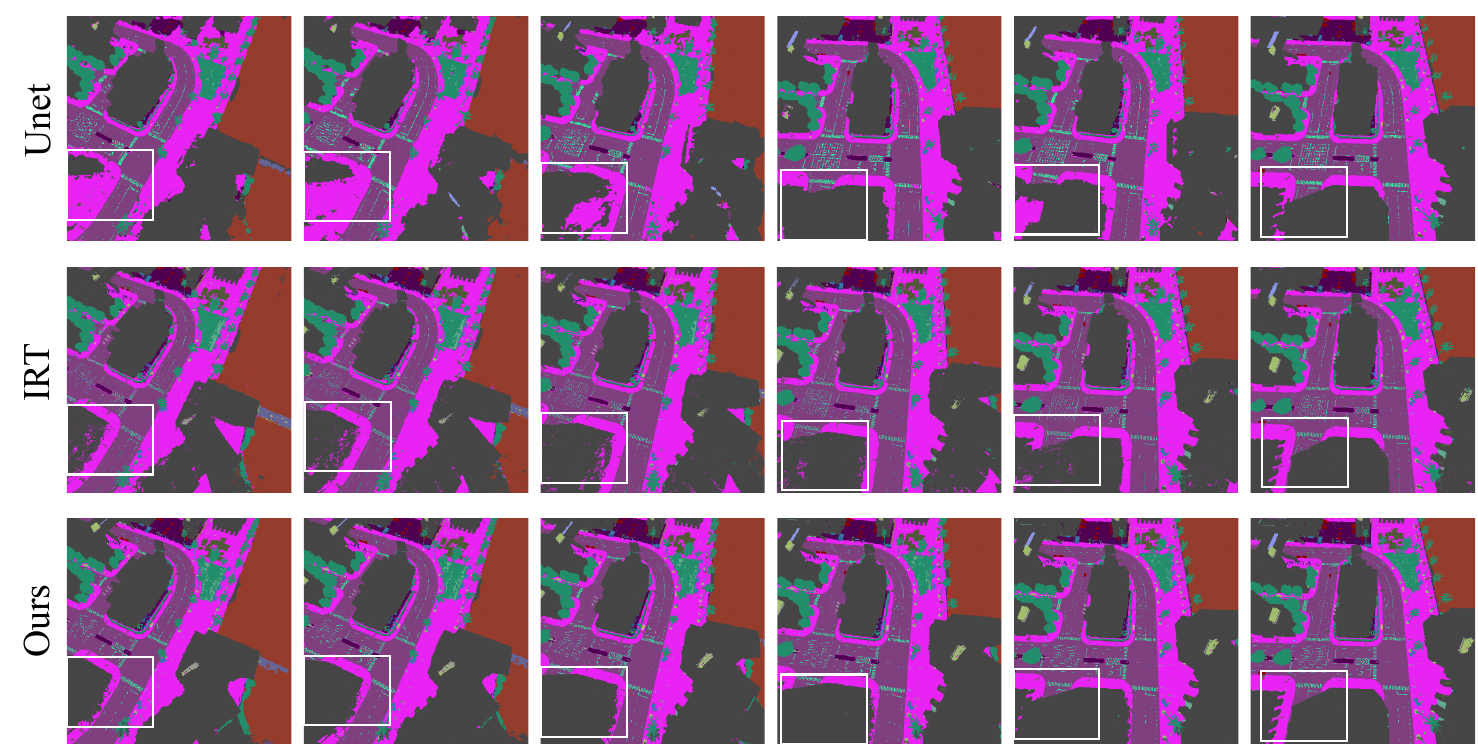}
\caption{The view-consistent results from sys \#4. The results show that our methods can generate results with more accuracy and view consistency. The labels in white boxes are not accurate or view-consistent using other methods.}
\label{view-consistency}
\end{figure*}

\textbf{Semantic Consistency Comparison}. We compare the segmentation view-consistency of different methods, including Unet, IRT, and our approach. As shown in Fig.~\ref{view-consistency}, Unet demonstrates low accuracy in classifying buildings from the initial viewpoints. Since it processes images from each viewpoint independently, it fails to maintain view consistency. In contrast, IRT employs an implicit neural network to construct a continuous spatial mapping function, while our method utilizes a point cloud to represent the scene's spatial information, resulting in improved view consistency. This improvement can be attributed to the introduction of aggregation losses and pseudo labels generated by SAM, which enhance the representation of boundary regions and preserve the structure of objects. For instance, our method produces more complete building structures. Compared to IRT, our approach not only ensures segmentation accuracy and view consistency but also achieves higher computational efficiency.

\begin{table*}[!t] 
\centering
\caption{Ablation study results.}
\label{tab:ablation_study}
\begin{tabular}{cccc|cccccccccc}
\toprule
Baseline & +Pseudo Labels & + $\mathcal{L}_a^{2D}$ & +$\mathcal{L}_a^{2D}$ & sys \#1 & sys \#2 & sys \#3 & sys \#4 & sys \#5 & sys \#6 & real \#1 & real \#2 & real \#3 & AVG \\\midrule
$\checkmark$ & & & & 58.51 & 29.11 & 56.96 & 60.62 & 41.35 & 50.18 & 58.96  & 84.91 & 76.46 &  57.45  \\ 
$\checkmark$ & $\checkmark$& & & 59.54 & 26.81 & 56.98 & 59.62 & 44.05 & 49.01  & 61.71 & 83.13  &  79.62 & 57.82 \\ 
$\checkmark$ &$\checkmark$& $\checkmark$ & &60.27 & 29.08 & 58.15 & 60.28 & 45.51  & 49.63 & 62.03 & 84.33 & 80.11  & 58.82  \\ 
$\checkmark$& $\checkmark$&$\checkmark$&$\checkmark$&\textbf{61.45} & \textbf{38.98} &\textbf{58.51} & \textbf{62.99} & \textbf{46.40} & \textbf{52.14} & \textbf{63.31} & \textbf{86.87} & \textbf{80.21}& \textbf{61.21}   \\

\bottomrule
\end{tabular}
\end{table*}

\begin{table*}[!t]
\centering
\label{combined-metric-results}

\begin{minipage}[b]{0.24\textwidth}
\label{tab:ratio}
\centering
\caption{The results of different ratios}
\begin{tabular}{c|ccc}
\toprule
Ratio & sys \#1 & real \#1 & AVG \\\midrule
1:7 & 59.10 & 61.32 & 60.21 \\
1:8 & \textbf{59.54}  & \textbf{61.71} & \textbf{60.63} \\
1:9 & 59.46  & 61.47 & 60.47 \\
\bottomrule
\end{tabular}
\end{minipage}%
\hspace{0.5cm}
\begin{minipage}[b]{0.24\textwidth}
\label{tab:2D}
\centering
\caption{The results of coefficients for $\mathcal{L}^{2D}_a$}
\begin{tabular}{c|ccc}
\toprule
Weight & sys \#1 & real \#1 & AVG \\\midrule
1.0  & 60.10 & \textbf{62.17} & 61.14 \\
0.5  & \textbf{60.27} & 62.03 & \textbf{61.15} \\
0.1 & 58.81 & 60.58 & 59.60 \\
\bottomrule
\end{tabular}
\end{minipage}%
\hspace{0.5cm}
\begin{minipage}[b]{0.24\textwidth}
\label{tab:3D}
\centering
\caption{The results of coefficients for $\mathcal{L}^{2D}_a$}
\begin{tabular}{c|ccc}
\toprule
Weight & sys \#1 & real \#1 & AVG \\\midrule
1.0  & 60.38 & 57.44 & 58.91 \\
0.5  & 61.05 & 59.94 & 60.50 \\
0.1 & \textbf{61.45} & \textbf{63.31} & \textbf{62.38} \\
\bottomrule
\end{tabular}
\end{minipage}%
\hspace{0.5cm}


\end{table*}

\subsection{Ablation Study}

In this subsection, we conduct ablation experiments to verify the following: (1) the necessity of pseudo labels generated by SAM2,
(2) the necessity of the 2D aggregation loss $\mathcal{L}_a^{2D}$, and (3) the necessity of the 3D aggregation loss $\mathcal{L}_a^{3D}$. We use the method that employs only basic semantic feature splatting as our baseline. The evaluation results are presented in the Table \ref{tab:ablation_study}.

First, we incorporate pseudo labels generated by SAM2 into the optimization process. Due to the weaker spatial continuity of the point cloud’s explicit spatial structure compared to implicit neural networks, our proposed method initially exhibited suboptimal performance in boundary regions. To address this, we use SAM2 to generate view-consistent pseudo labels as additional supervisions specifically for these regions. As shown in Table \ref{tab:ablation_study}, this strategy proves effective across almost all sub-datasets. With pseudo labels, the method outperform baseline over 0.4\%
\begin{figure*}
    \centering
    \includegraphics[width=1.0\linewidth]{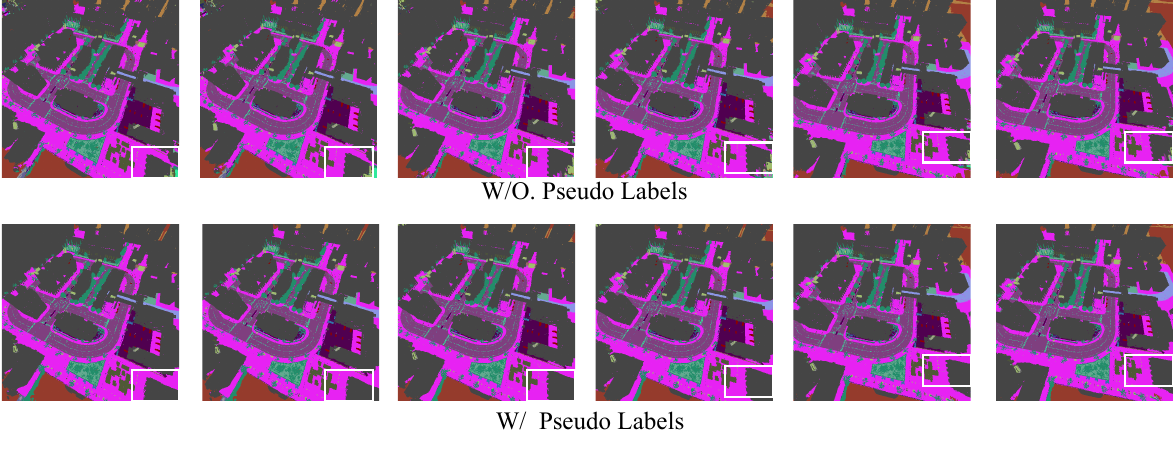}
    \caption{The pseudo labels for boundary regions assist the network in improving accuracy in these challenging areas (shown in white boxes). }
    \label{fig:pseudo}
\end{figure*}
Figure \ref{fig:pseudo} also presents the visual results, clearly demonstrating the improved effectiveness in boundary regions, particularly in the buildings within the sys \#1 sub-dataset, better to zoomed in.

Secondly, we conduct experiments to validate the necessity of the two aggregation losses. We gradually incorporate $\mathcal{L}_a^{2D}$ and $\mathcal{L}_a^{3D}$ into the optimization process. The $\mathcal{L}_a^{2D}$ loss encourages semantic features within the same class to become more similar, while $\mathcal{L}_a^{3D}$ ensures that semantic features in neighboring spatial regions are also similar. Notably, due to the splatting strategy, $\mathcal{L}_a^{3D}$ indirectly influences the semantic feature map. As shown in Table \ref{tab:ablation_study}, incorporating $\mathcal{L}_a^{2D}$ improves performance by 1.1\%, and further adding $\mathcal{L}_a^{3D}$ increases performance by an additional 2.4\%, demonstrating the effectiveness of these losses in enhancing segmentation quality.
\begin{figure*}
    \centering
    \includegraphics[width=1.0\linewidth]{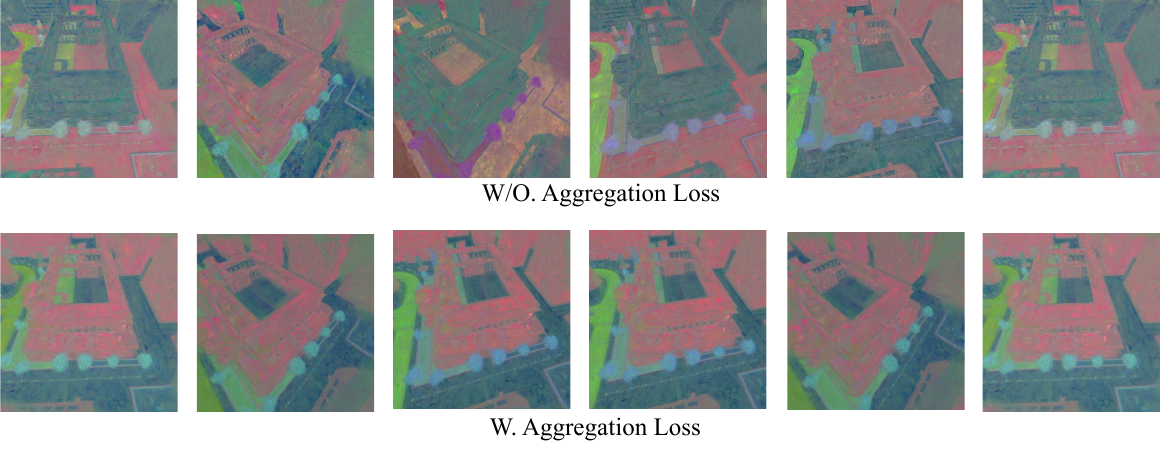}
    \caption{We calculate the PCA results of splatted semantic feature map. The results show the aggregation losses help the network enhance the stability of semantic features.}
    \label{fig:feature_maps}
\end{figure*}
We apply Principal Component Analysis (PCA) to reduce the dimensions of the splatted semantic features from 16 to 3, enabling the generation of RGB visual representations, as shown in Figure \ref{fig:feature_maps}. From the results on the sys \#6 sub-dataset, we observe a significant improvement in the similarity of features within the same class and the consistency of features across different viewing angles.

\subsection{Parameter Study}
We observe that the hyper-parameters, including the ratio of two types views(with ground truth and with pseudo label) and the weights of $\mathcal{L}_a^{2D}$ and $\mathcal{L}_a^{3D}$, play a crucial role in the optimization process of our method. To determine the optimal values for these parameters, we conduct experiments on both a synthetic sub-dataset (sys \#1) and a real sub-dataset (real \#1).

\textbf{The ratio of two types views.}
The pseudo labels generated by SAM2 contain some inaccuracies and are not sufficiently precise for semantic segmentation. Furthermore, the proportion of pseudo-labeled data significantly outweighs the ground truth data, e.g., 97\% pseudo labels versus 3\% ground truth. Selecting views with ground truth and pseudo labels with equal probability would lead to imbalanced supervision, resulting in suboptimal performance. To mitigate the negative impact of pseudo labels, we experimented with different ratios of selected views. An insufficient proportion of pseudo labels results in inadequate supervision of boundary areas, while an excessive proportion introduces inaccuracies and unrefined labels, negatively impacting the final results. As shown in Table \ref{tab:ratio}, our method achieves the best performance on both SYS 1 and REAL 1 when the ratio of views with ground truth to views with pseudo labels is set to $1:8$.

\textbf{The weight of $\mathcal{L}_a^{2D}$.}
The aggregation loss $\mathcal{L}_a^{2D}$ is pivotal in enhancing feature similarity within the same class at the level of the 2D semantic feature map. By enforcing semantic similarity among neighboring feature points, it promotes a more compact and cohesive feature distribution for points belonging to the same class. This not only boosts the segmentation model's discriminative power but also ensures improved semantic continuity across segmented regions, especially at class boundaries. Such continuity effectively reduces misclassification errors and boundary inconsistencies, leading to cleaner and more accurate segmentation results. Experimental results, as presented in Table \ref{tab:2D}, confirm that carefully tuning the weight coefficient $\mathcal{L}_a^{2D}$ yields the best trade-off between feature consistency and segmentation performance. These findings underscore the critical role of optimizing the $\mathcal{L}_a^{2D}$ coefficient to achieve superior overall performance. The best coefficient of $\mathcal{L}_a^{2D}$ is 0.5.

\textbf{The weight of $\mathcal{L}_a^{3D}$.}
Similar to $\mathcal{L}_a^{2D}$, the aggregation loss $\mathcal{L}_a^{3D}$ is designed to enhance feature similarity, but at the 3D spatial point level. By enforcing semantic similarity among spatially adjacent points in the 3D point cloud, $\mathcal{L}_a^{3D}$ ensures that points belonging to the same class exhibit consistent features across the spatial domain. This improves spatial coherence in segmentation and strengthens the alignment between the 3D structure of the point cloud and its semantic representation. The weight coefficient $\mathcal{L}_a^{3D}$ controls the relative impact of $\mathcal{L}_a^{3D}$ during the optimization process. As shown in Table \ref{tab:3D}, experiments evaluating different values of $\mathcal{L}_a^{3D}$ reveal its significant influence on segmentation quality. While $\mathcal{L}_a^{2D}$ focuses on local feature similarity within 2D feature maps, $\mathcal{L}_a^{3D}$ extends this consistency to the 3D spatial level. Their combined application ensures both precise local segmentation in 2D and robust spatial consistency in 3D. The results in Table \ref{tab:ablation_study} demonstrate that leveraging these two losses together leads to substantial improvements in segmentation accuracy and multi-view consistency. The best coefficient of $\mathcal{L}_a^{2D}$ is 0.1.

\section{Further Work.}
In this paper, our method demonstrates the efficiency and accuracy of splatting technology in multi-view image segmentation for remote sensing scenes. Moving forward, we plan to explore more appropriate 3D structure representations for sparse spatial characteristics, such as using linear functions instead of Gaussian functions. Additionally, we aim to integrate implicit neural functions with splatting technology to estimate a semantic neural function for each point in the point cloud, further improving spatial consistency.

\section{Conclusion}\label{section:conclusion}

In this paper, we aim to improve multi-view image segmentation for remote sensing of target scenes, especially under conditions with sparse semantic labels. We propose an efficient semantic splatting method based on Gaussian Splatting to address this challenge. To overcome the lack of labels in boundary regions, we introduce SAM2, which generates pseudo-labels for these areas. Furthermore, we design two aggregation losses: one at the 2D feature map level and another at the 3D spatial level, to enhance the spatial continuity of our method and boost its performance. Experimental results demonstrate that our approach achieves significant improvements in segmentation accuracy and low latency, providing a solid foundation for practical applications.

\ifCLASSOPTIONcaptionsoff
  \newpage
\fi

\bibliographystyle{IEEEtran}
\bibliography{refbib}

\end{document}